\documentclass{article}

\usepackage[numbers]{natbib}

\usepackage{arxiv}

\usepackage[utf8]{inputenc} 
\usepackage[T1]{fontenc}    
\usepackage{hyperref}       
\usepackage{url}            
\usepackage{booktabs}       
\usepackage{amsfonts}       
\usepackage{nicefrac}       
\usepackage{microtype}      
\usepackage{lipsum}

\usepackage{graphicx} 
\usepackage{float}  
\usepackage{subcaption}

\usepackage{amsmath}
\usepackage{amssymb}

\title{On the Vector Space in \\ Photoplethysmography Imaging}

\author{
  Christian S. Pilz\thanks{http://www.cancontrols.com} \\
 CanControls GmbH Aachen, Germany \\
  \texttt{pilz@cancontrols.com}
  \And
  Vladimir Blazek\thanks{http://www.medit.hia.rwth-aachen.de} \\
  RWTH Aachen, Germany \\
  \texttt{blazek@hia.rwth-aachen.de}
  \And
  Steffen Leonhardt\thanks{http://www.medit.hia.rwth-aachen.de} \\
  RWTH Aachen, Germany \\
  \texttt{leonhardt@hia.rwth-aachen.de}
}

\begin{document}
\maketitle

\begin{abstract}
We study the vector space of visible wavelength intensities from face videos widely used as input features in Photoplethysmography Imaging (PPGI). Based upon theoretical principles of Group invariance in the Euclidean space we derive a change of the topology where the corresponding distance between successive measurements is defined as geodesic on a Riemannian manifold. This lower dimensional embedding of the sensor signal unifies the invariance properties with respect to translation of the features as discussed by several former approaches. The resulting operator acts implicit on the feature space without requiring any kind of prior knowledge and does not need parameter tuning. The resulting feature’s time varying quasi-periodic shaping naturally occurs in form of the canonical state space representation according to the known Diffusion process of blood volume changes. The computational complexity is low and the implementation becomes fairly simple. During experiments the operator achieved robust and competitive estimation performance of heart rate from face videos on two public databases.
\end{abstract}

\keywords{Photoplethysmography Imaging \and Computer Vision \and Linear Algebra \and Group Theory \and Manifolds}

\begin{figure}[H]
\begin{center}
\includegraphics[scale=.3]{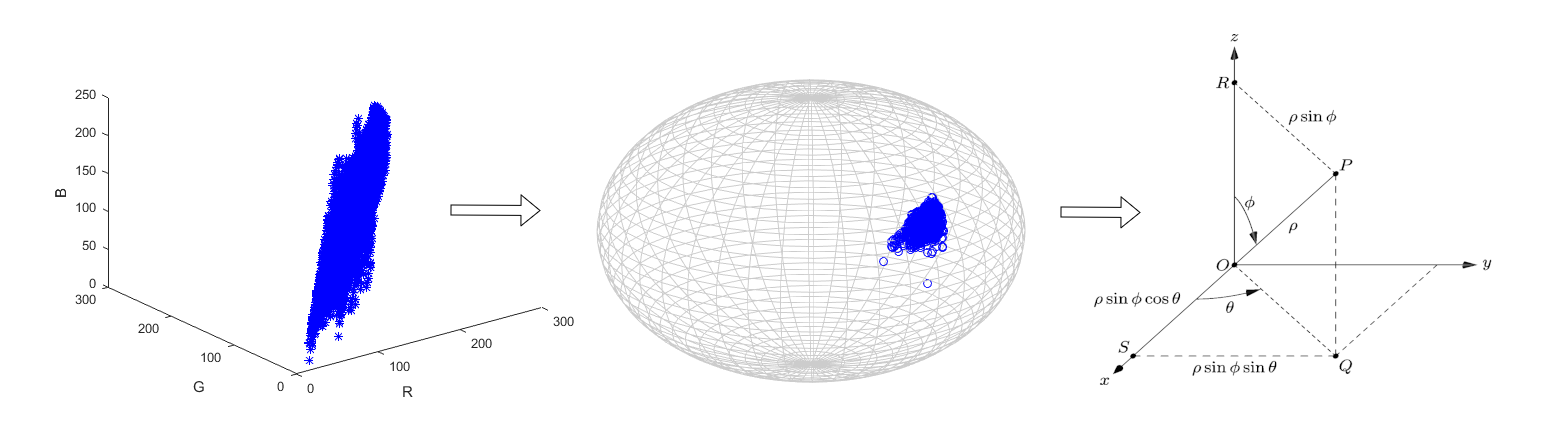}
\caption{The PPGI Signal as distribution of pixel intensities in the Euclidean RGB space is embedded in a lower dimensional space on the unit sphere. On the left, the skin pixel intensities are drawn in the RGB coordinate system. In the middle, the pixel intensities are mapped onto the unit sphere. On the right, their schematic representation as spherical coordinates.}
\label{fig:0}       
\end{center}
\end{figure}

\newpage

\section{Introduction}
	\label{sec:introduction}
	
Nearly 70 years ago, short after the end of world war II in 1948, Norbert Wiener published, his sociopolitical often controversial discussed work, \textit{Cybernetics or Control and Communication in the Animal and the Machine} \cite{Wiener1948}. During this period and short after, mankind already discovered most of the important principles found today in everyday technology. Although we have not seen yet a fully functional realization of Wieners visions of self-regulating mechanisms, but we are able to trace ongoing and fast emerging progress in the computational interpretation of sensors, signals and systems which yields at least to a direction of a soup\c{c}on of artifical intelligenz.\newline
\newline
As part of natural social interaction the human face with it's contingent of verbal activity, non-verbal behaviour as well as it's appearance is reflecting the majority of interpretable signal sources by computer systems. In contrast to the non-verbal behavioral signals the facial appearance is able to provide changes in peripheral nervous system as well as central nervous system surrogate states by the analysis of skin blood perfusion. Though the tiny intensity changes can not be perceived by human eyes. However with the help of opto-electronic circuitry this information has become accessible. Scientifically a plain disposability is nothing special. Under the context of non-obtrusive remote measurability this new technology enables acquisitions of biological and cognitive human data under exceptional situations potentially leading to new findings and application fields. Basically, nearly every tasks targeting on the specific dependent variables, formally captured by cable mounted sensors sticked onto human skin, can be sensed by the camera based counterpart as well. The possible range of new applications and analysis legitimates to name its potential as ground breaking. The topic is traditional anchored in the medical sciences. However, currently the focus elementary changed into direction of computer vision. Here the role of physiological states has a large impact. It is primarily used during human state computing tasks, where the radiation unnoticeable transports information from face without contact holding states of affective nature. \newline
\newline
During the last years measuring blood volume changes and heart rate measurements from facial images gained attention at top computer vision conferences \cite{Li2014,Osman2015,Tulyakov2016,Pilz2017,Pilz2018,Chen2018} frequently.
Most of these contributions focus on how to cope with motion like head pose variations and facial expressions since any kind of motion on a specific skin region of interest will destroy the raw signal in a way that no reliable information can be extracted anymore. Beside from being able to estimate vitality parameters like heart rate and respiration, the functional survey of wounds as well as quantification of allergic skin reaction \cite{Blazek2005} are further topics of discovered employments of skin blood perfusion analysis. Recently, prediction of emotional states, stress \cite{McDuff2014,Ramirez2014,Blazek2017}, fatigue \cite{Sundelin2013} and sickness \cite{Henderson2017} became interesting new achievements in this area, pushing the focus of this technology further towards human-machine interaction.\newline
\newline
In contrast to the genuine medical use-case of the technology, in computer vision and human-machine interaction we can't expect any cooperative behavior of the user without introducing lack of convenience and a reduction of the general user acceptance. Further, beyond any well tempered clinical and laboratory like scenarios, the majority application will face strong challenging environmental changes and differences much more quite common. Thus, there's an emerging demand to produce better features and models significant more robust to nuisance factors, still preserving the desired target information. To reach such a formulation a fundamental profound understanding of the underlying optical and mathematical properties is one of the current foci of this research discipline.\newline
\newline
The main contribution of this work is a mathematically analysis of the PPGI's feature space determined over facial skin pixels. The general aim is to study the properties and behaviour of the features with respect to the influence of the actions of the Group acting on this space when induced by natural head and face motion. As result a new feature operator is developed and evaluated against common operators on various data sets. These sets are collected to explicitly study the influence of typical nuisance factors. To support an efficient dissemination and to speed up the research progress in this field we have encapsulated all feature operators and the reference implementation of the stochastic model of blood volume changes into a new object-orientated MATLAB toolbox. The code is public available under \url{http://bit.ly/PPGI-Toolbox}. The toolbox provides the necessary code to reproduce all results presented in this work.\newline
\newline
The outline of this work is as follows. From the historical genuine up to the development of the state of the art in computer vision and bio-medical engineering, the methodology of heart rate estimation from face videos will be reviewed. Followed by theoretical aspects, the feature space will be analyzed and the proposed methodology mathematically described. Based upon an extensive evaluation on different databases the results will be presented and finally discussed.

	\section{Related Work}
    \label{sec:related_work}
    
The historical genuine of the term Photoplethysmography, short PPG, dates back to the late first half of the 20th century, when the two scientists Molitor and Kniazak \cite{MolitorKniazak1936} recorded peripheral circulatory changes in animals. A year later, Hertzman \cite{Hertzman1937} introduced the term Photoelectric Plethysmograph as "the amplitude of volume pulse as a measure of the blood supply of the skin". Hertzman's instrumentation comprised mainly of a tungsten arc lamp and a photomultiplier tube. During the same time the methodology was described by a German scientist in a medical journal \cite{Mathes1935}. However, today it is not reproducible anymore who really invented this technology. In literature Hertzman is accredited with this reputation usually.\newline
\newline
Around 50 years later the advancement to the classical PPG, the camera based PPGI (with I for Imaging) method, was introduced by the pioneering work of Blazek \cite{Blazek1985}. The basic principle behind the measurement of blood volume changes in the skin by means of PPGI (as well as PGG too) is the fact that the oxygen binding ferrous protein complex hemoglobin in the blood absorbs specific frequency bands of light many times more strongly than the remaining skin tissues. Accordingly, tiny intensity changes can be observed over specific frequency bands (e.q. the density of spectral lines of the emission spectra of iron) as oscillation caused by the quasi periodic rhythm of the human heart. In PGG a part of the skin surface is illuminated by dedicated light sources like illumination panels consisting of LED. In PPGI a common CCD camera is used as detector and the illumination can be as well as a common ambient light for which the intensity of backscattered optical radiation, eq. reflected light, is calculated \cite{Allen2007,Huelsbusch2008,Verkruysse2008}. 
\newline
\newline
In general the computational pipeline to determine vitality parameters and its derivatives from blood volume changes can be regarded as classical signal processing chain. Typically, from a skin region of interest (ROI) features are calculated, filtered and analyzed by spectral methods \cite{Huelsbusch2008,Poh2010,Verkruysse2008}. The first published visualization of pulsatile skin perfusion patterns in the time and frequency domain is given by Blazek \cite{Blazek1985}. However motion of the skin ROI \cite{Huelsbusch2008} and micro motion of the head due to cardiac activity \cite{Blanik2014,Moco2015} inherently induces artifacts into the extracted signal, especially when lighting is neither uniform nor orthogonal \cite{Moco2016}. Canceling motion artifacts during signal processing became one of the most important aspect for reliable skin blood perfusion measurements. An early idea of skin ROI  motion compensating is to track every skin pixel position by optical flow methods directly in the image plane \cite{Huelsbusch2008}. However this doesn't account for any change of illumination. Poh \textit{et al.} \cite{Poh2010} proposed to extract motion components in the signal by blind source separation using Independent Component Analysis (ICA) over the different color channels. Wedekind \textit{et al.} \cite{Wedekind2017} compared ICA in multiple setting and Principal Component Analysis and showed limitations of either transform. Further, the ICs cannot be obtained in a deterministic order \cite{Cardoso1999}. A solution to this problem is discussed by Macwan \textit{et al.} \cite{Macwan2018}. Tarassenko \textit{et al.} \cite{Tarassenko2014} tried  to cope with light flicker by using an auto-regressive modeling and pole cancellation. De Haan and Jeanne \cite{HaanJeanne2013} and De Haan and Van Leest \cite{HaanLees2014} proposed to map the PPGI-signals by linear combination of RGB data to a direction that is orthogonal to motion induced artifacts. An alternative approach, which does not require skin-tone or pulse-related priors in contrast to the channel mapping algorithms, determines the spatial subspace of skin-pixels and measure its temporal rotation for signal extraction \cite{Wang2015}. Tulyakov \textit{et al.} \cite{Tulyakov2016} proposed matrix completion to jointly estimate reliable regions and heart rate estimates whereby Li \textit{et al.} \cite{Li2014} applied an adaptive least square approach to extract robust pulse frequencies. Both reported  performance gains similar to De Haan and Jeanne \cite{HaanJeanne2013}. Interestingly they used the often criticized compressed videos of the MAHNOB-HCI database \cite{Soleymani2012} during their experiments. This leaves reasonable doubts on the validity of results since it is well known that any kind of image compression will destroy the underlying tiny perfusion signal \cite{McDuff2017}. Wang \textit{et al.} \cite{Wang2017} reported an orthogonal behavior of skin color and motion artifacts derived by optical properties but introduced a static projection operator for feature transformation and represented their results on private data.  An entirely different model was introduced by Pilz \textit{et al.} \cite{Pilz2017}. Here, the quasi-periodic nature of the blood volume changes is modeled as stochastic resonator based upon a diffusion process. A group theoretic deviated feature transform for motion compensation is introduced by Pilz \textit{et al.} \cite{Pilz2018}. Following the popularity of Deep Learning Chen and McDuff \cite{Chen2018} claim to outperform recent algorithms using a convolutional network architecture (CNN) for modelling motion representation. However, they also reported some findings on the compressed videos of the MAHNOB-HCI database and they didn't provide their CNN implementation or at least the trained model yet.

\newpage

\section{Methodology}
\label{sec:methodology}

\begin{figure}[h]
\begin{center}
\includegraphics[scale=.3]{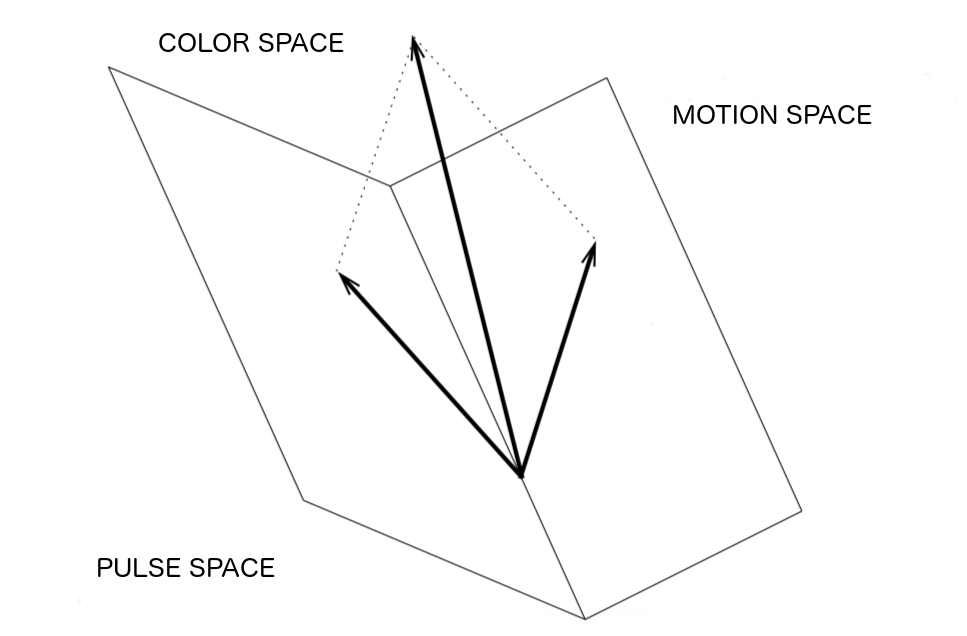}
\caption{The actual state-of-the-art paradigm in understanding major PPGI signal components in the feature space. With $\vec{c}=\vec{p}+\vec{m}$, $\vec{p}$ and $\vec{m}$ are two orthogonal vectors with $\vec{p} \cdot \vec{m}^T=0$.}
\label{fig:1}       
\end{center}
\end{figure}

Meanwhile, it is well understood that subject motion and fast strong changes of illumination alters the distribution of pixel intensity negatively making it quite difficult to extract skin perfusion signals from video images. It is assumed that the perfusion signals exits in either case and its further assumed that it is combined together with the distribution of intensity belonging to motion forces by some unknown operator. Following the basic principles of the Hilbert projection theorem. If $\mathcal{V}$ is a closed subspace of the Hilbert Space $\mathcal{H}$ and $x \in \mathcal{H}$, then there is a unique element $\hat{x} \in \mathcal{V}$ such that
\begin{equation}
\| x-\hat{x} \| =  \underset{y\in \mathcal{V}}{\textrm{inf }} \| x-y \| 
\end{equation}
and only if  $\hat{x} \in \mathcal{V} $ and $(x-\hat{x}) \in \mathcal{V}^{\perp}$ where $\mathcal{V}^{\perp}$ is the orthogonal complement to $\mathcal{V}$ in$\mathcal{H}$.
Then, the current paradigm in understanding PPGI signal components in the feature space assumes that blood volume changes exist in a lower dimensional space where this space is orthogonal to any kind of motion induces signal components. Derived from optical properties by De Haan \textit{et al.} and Wang \textit{et al.} \cite{HaanJeanne2013,Wang2017} and from Group theoretic principals by Pilz \textit{et al.} \cite{Pilz2018} this can be expressed as $\vec{c}=\vec{p}+\vec{m}$. $\vec{p}$ and $\vec{m}$ are two orthogonal vectors with $\vec{p} \cdot \vec{m}^T=0$, thus statistically linear independent. This principal is illustrate in Figure ~\ref{fig:1}. In the following we explain the Group theoretic principals behind motion robust sensing of blood volume changes as introduced by Pilz \textit{et al.} \cite{Pilz2018}. Based upon the analysis of the properties of the resulting linear operator we demonstrate that there exists an equivalent implicit operator. This operator maps the observation onto an embedded Riemannian submanifold of the Euclidean space. And we show that the corresponding directional statistics evolve in form of the previously published Diffusion process model as function of time \cite{Pilz2017}.

\subsection{Basic Principals of Group Invariance}

Consider a finite topological group
\begin{equation}
\mathcal{G}=\{G_1,...,G_M\}
\end{equation}
of $M$ distinct actions on a topological space
\begin{equation}
\mathbb{X},G_i : \mathcal{X}\rightarrow \mathcal{X}.
\end{equation}
A real valued function $f\left(x\right)$ on $\mathcal{X}$ is said to be invariant under $\mathcal{G}$ if
\begin{equation}
f\left(Gx\right)=f\left(x\right) \textrm{for } G \in \mathcal{G}
\end{equation}
\newline
Regarding a common optical sensor signal $\{\vec{p}_i: i=1,...,m\}$
\begin{equation}
\vec{p} \in  \mathbb{R}^n = \{Red,Green,Blue\}, n=3
\end{equation}
as spatial expectation over a skin operator $s$ and function of time $t$
\begin{equation}
\vec{x}(t)=\int_{0}^{\infty} \mathbb{E}[\{ \vec{p_i} \mid s(\vec{p_i}) \}] \mathrm{d}t
\end{equation}
we assume this multivariate observation is drawn by a normal distribution
\begin{equation}
\vec{x}(t) \sim \mathcal{N} (\mu,\sigma^2).
\end{equation}
\newline
Local invariance of blood volume changes as function of time for each input feature $\vec{x}(t)$ under transformations of a differentiable local group of local transformations $\mathcal{L}_T$ \cite{Schoelkopf1998}
\begin{equation} \label{eq:1}
\frac{\partial }{\partial T}\big\vert_{T=0}=f(\mathcal{L}_T,\vec{x}(t))=0
\end{equation}
can be approximately enforced by minimizing the regularizer
\begin{equation} \label{eq:2}
\frac{1}{l}\sum_{j=1}^l(\frac{\partial }{\partial T}\big\vert_{T=0}f(\mathcal{L}_T,x_j))^2.
\end{equation}
For the covariance matrix of the observation $\{x_i: i=1,...,l\}$
with respect to the transformations $\mathcal{L}_T$
\begin{equation}
C:=\frac{1}{l}\sum_{j=1}^l(\frac{\partial }{\partial T}\big\vert_{T=0}\mathcal{L}_T,x_j)(\frac{\partial }{\partial T}\big\vert_{T=0}\mathcal{L}_T,x_j)^\top
\end{equation}
and the corresponding symmetric eigenvalue problem
\begin{equation}
CV=V\Lambda
\end{equation}
we find an operator $P$ with corank $k=1$ for
\begin{equation}
\lim\limits_{l \to \infty}P=I-VV^\top
\end{equation}
and the corresponding feature vector
\begin{equation}
\tilde{x}=P\cdot x.
\end{equation}
\newline
The observation $\{\tilde{x_i}: i=1,...,l\}$
is defining the null space of the projection operator $P$
\begin{equation}
H_P=N(P)
\end{equation}

\subsection{The embedded Riemannian submanifold}

In general, we're specially interested in the properties of the projection operator $P$ and the resulting linear subspace $H_P$ since the direction of $V$ is assumed to carry most of the PPGIs motion signal component and the subspace $H_P$ the desired quasi periodic perfusion component. It is a well-known from the theory of Banach algebras that the spectral radius $\rho$ of any  $A\in \mathcal{M}_n$ is given by Gelfand’s formula
\begin{equation}
\rho(A)=\lim\limits_{n \rightarrow \infty}{\left\|{A^n}\right\|^{1/n}}
\end{equation}
for any matrix norm $\|\cdot\|$ on $\mathcal{M}_n$.
For a matrix, the spectrum is just the collection of eigenvalues with $\rho(A):=max\{\left|\lambda\right|,\lambda \textrm{ eigenvalue of } A\}$. For the projection operator $P$, where $V$ is represented by the eigenvector with the largest eigenvalue, this implies a reduction of the spectral radius of the observation $X$
\begin{equation}
\rho(\tilde{X})<\rho(X)
\end{equation}
The projection $P$ removes the direction of the largest variance and puts more emphasis on the directions which vary less. It should be clear that the computation of the projection is an explicit operator which has to be estimated on an observation $\{x_i: i=1,...,l\}$. A more convenient way of incorporating invariance to the feature space would be to define an implicit operator. \newline
\newline
Let $\lambda$ be an eigenvalue of $A$, and let $y\neq 0 $ be a corresponding eigenvector. From $Ay=\lambda y$, we have $AY=\lambda Y$ where $Y:=\left[ y\mid ... \mid y \right] \in \mathcal{M}_n \setminus  \{0\}.$ It follows $\left|\lambda \right| \|X\| = \|\lambda Y\| = \|AY\| \leq \|A\|\|Y\|$, and simplifying by $\|Y\|\left(>0\right)$ gives $\left|\lambda \right| \leq \|Y\|$. Taking the maximum over all eigenvalues $\lambda$ results in $\rho(A) \leq  \|Y\|$.
\newline
Now, regarding the optical sensor signal $\vec{p}_i$ relative to the spectral radius of its elements $\rho(\vec{p})$ results in
\begin{equation}
\vec{x}_{\vec{1}}(t)=\int_{0}^{\infty} \mathbb{E}[\{ \frac{\vec{p_i}}{\|\vec{p}\|} \mid s(\vec{p_i}) \}] \mathrm{d}t
\end{equation}
with 
\begin{equation}
\rho(X_{\vec{1}})<\rho(X)
\end{equation}
and distributed on $\mathbb{S}^2=\{ \vec{x} \in \mathbb{R}^3 \mid \|\vec{x}\|=1 \}$ the unit sphere as embedded Riemannian submanifold of the Euclidean space $R^3$.\newline
\newline
Intuitively, for real valued observations $\{y_i: i=1,...,m \in  \mathbb{R}\}$ the mean is given by $\mu=\frac{1}{m}\sum_{i=1}^m {y_i}$. However, for the more general settings $\{y_i: i=1,...,m \in  \mathcal{M}\}$ we do not have the possibility to find such a closed form solution. The solution has to be solved for the optimization problem
\begin{equation}
\mu \in \underset{x\in \mathcal{M}}{\arg\min} \frac{1}{m}\sum_{i=1}^m d^2_{\mathcal{M}}\left(x,x_i \right)=\underset{x\in \mathcal{M}}{\arg\min}F\left(x\right).
\end{equation}
by gradient descent algorithm. The resulting $\mu$ is called Riemannian center of mass or Karcher mean \cite{Karcher1977}. The optimality conditions is give by
\begin{equation}
0\overset{!}{=} \frac{2}{m}\sum_{i=1}^n log_{x^*}x_{i}=\nabla F\left(x^*\right) \in T_{x^*}\cal{M}
\end{equation}

Considering the embedded dimensions of the unit sphere given by its spherical coordinates 
\begin{equation}
r=\sqrt{x^2+y^2+z^2}=1
\end{equation}
\begin{equation}
\theta=arccos\frac{z}{\sqrt{x^2+y^2+z^2}}=arccos \frac{z}{r}=arccos{z}
\end{equation}
\begin{equation}
\varphi=arctan \frac{y}{x}
\end{equation}
the directional statistics are represented by the von Mises-Fisher distribution
\begin{equation}
p\left(x;\mu,\kappa\right):=c_{p}\left(\kappa\right)e^{\kappa\mu^{T}x}
\end{equation}
with the normalization constant given by
\begin{equation}
c_{p}\left(\kappa\right)=\frac{\kappa^{p\slash2-1}}{\left(2\pi\right)^{p\slash2}I_{p\slash2-1}\left(\kappa\right)}
\end{equation}
where $I_s\left(\kappa\right)$ denotes the modified Bessel function of the first kind.
The spherical direction $\varphi$ evolves according to the principals of the stochastic differential equation given by
\begin{equation}
\frac{d^2c_n(t)}{dt^2}=-(2\pi{nf(t)})^2{c_n(t)}+e_n(t)
\end{equation}
whereby $\theta$ can be expressed as part of a Wiener process given by
\begin{equation}
\frac{\mathrm{d}^2\theta(t)}{\mathrm{d}t^2}=w(t).
\end{equation}
For a detailed discussion on the Diffusion process see the previous works of Pilz \textit{et al.} \cite{Pilz2017}.

\newpage

\section{Experiments}
\label{sec:fexperiments}

The evaluation of the described feature operator was conducted by experiments against different methods on two public available databases. We decided to compare the operator against the baseline green channel expectation \cite{Huelsbusch2008,Verkruysse2008}, the Spatial Subspace Rotation (SSR) \cite{Wang2015}, the Projection Orthogonal to Skin (POS) \cite{Wang2017} and the Local Group Invariance (LGI) \cite{Pilz2018} method. All experiments were executed on the French UBFC-RPPG \cite{Bobbia2017} and the German LGI Multi-Session face video database \cite{Pilz2018}. In the following Figures ~\ref{fig:ubfc_examples} and ~\ref{fig:lgi_examples} some representative images of these two databases are illustrated. The UBFC and the LGI 
\begin{figure} [!]
\begin{center}
  \begin{subfigure}[h]{0.8\textwidth}
    \includegraphics[width=\textwidth]{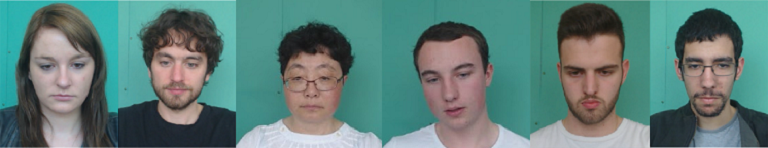}
    \caption{UBFC-RPPG}
    \label{fig:ubfc_examples}
  \end{subfigure}
  \begin{subfigure}[h]{0.8\textwidth}
    \includegraphics[width=\textwidth]{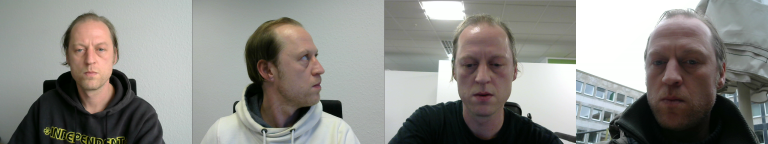}
    \caption{LGI Multi-Session}
    \label{fig:lgi_examples}
  \end{subfigure}
  \caption{Example images taken from the UBFC-RPPG and LGI Multi-Session database.}
  \end{center}
\end{figure}
database were created using a custom C++ application for video acquisition with a simple low cost Logitech webcam and a CMS50E transmissive pulse oximeter to obtain the ground truth PPG data comprising of the PPG waveform. The total amount of face video recordings yields to 150 sequences containing several couple of minutes respectively. During the recordings, the subjects of the UBFC trials performed moderate face and head motions under indoor environments whereby the LGI recordings contain scenarios from resting and head rotation over sport activities to natural outdoor conversations. Therefore, the evaluation concept ranges from cooperative to challenging scenarios which should be become noticeable in form of the prediction accuracy results.\newline
The primary signal processing procedure is selected to be equal for every approach and database. For each video frame a common Viola-Jones face finder was used to pre-select the region of interest. A simple skin operator was applied onto the region by thresholding the blue- and red-difference chroma components. For the set of obtained RGB-pixels the different approach specific operators were computed and stored as time series for further spectral processing and interpretation. Each signal obtained by the different algorithms was band-filtered in the range between 0.5 and 2.5 Hz. All filtered signals were then analyzed by standard Fourier based spectral method with windows size of 256 samples and overlap of 90 percent. A maximum peak energy criterion was applied over the spectral traces to determine the heart rate. All PPG ground truth signals were analyzed in the same way but initially resampled to the camera frame rate. Correlation coefficients were computed against the PPG reference heart rate together with the root-mean-square error (RMSE) for each user and algorithm respectively. We did not perform a Signal-to-noise ratio (SNR) comparison as proposed by De Haan and Jeanne \cite{HaanJeanne2013} and often used by several other authors \cite{HaanLees2014,Wang2017,Chen2018}. This might be useful on short video sequences with a more or less stationary frequency behavior of blood volume changes. A prospective consideration of a SNR metric for system evaluation should at least include information about it's variance computed on short term spectra.\newline
Table ~\ref{tab:table} represents the overview of heart rate prediction accuracy of the different operators on the different data sets. Compared to the baseline green channel, SSR and POS approach the LGI and the proposed spherical operator (SPH) are more robust especially under motion scenarios. Although the SPH operator cannot outperform the LGI approach in most cases, it's accuracy is operating in a very similar range. However, in case of fully uncontrolled scenarios with changing illumination as well as different head and face motion, as given in the LGI city talk session, no algorithm is able to perform reasonable well.\newline
In Figure 4 and 5 the box plot statistics for the UBFC and the LGI database are visualized. In addition to Table ~\ref{tab:table} the influence of the Diffusion process incorporated into the  LGI and SPH approach is constituted. Both approaches benefit from its stochastic interpretation as quasi-periodic nature of blood volume changes.

\begin{table}[h]
 \caption{Comparison of heart rate prediction accuracy utilizing different feature operators on diverse face video databases. Each corresponding table entry represents the average Pearson's correlation coefficient together with the average root-mean-square error (RMSE) value in BPM. \newline}
  \centering 
  \begin{tabular}{llllll}
    Database     & Green  & SSR  & POS & LGI & SPH\\
    \hline
    \hline
    UBFC & 0.16/22.1   & 0.54/4.95 & 0.68/4.42 & 0.75/5.94 & 0.73/3.21    \\
    LGI Resting     & 0.41/2.61 & 0.49/1.99  & 0.41/2.10 & 0.69/1.41 & 0.71/1.49    \\
    LGI Rotation     & 0.15/13.2 & 0.06/10.9  & 0.12/5.32 & 0.67/1.92 & 0.56/2.54    \\
    LGI Gym     & 0.01/33.5 & 0.03/21.2  & 0.15/12.2 & 0.42/2.65 & 0.26/3.65    \\
    LGI Talk     & 0.15/46.6 & 0.12/27.8  & 0.01/37.7 & 0.51/14.7 & 0.23/27.8    \\
    \hline
  \end{tabular}
  \label{tab:table}
\end{table}

\begin{figure} [h]
\begin{center}
  \begin{subfigure}[h]{0.4\textwidth}
    \includegraphics[width=\textwidth]{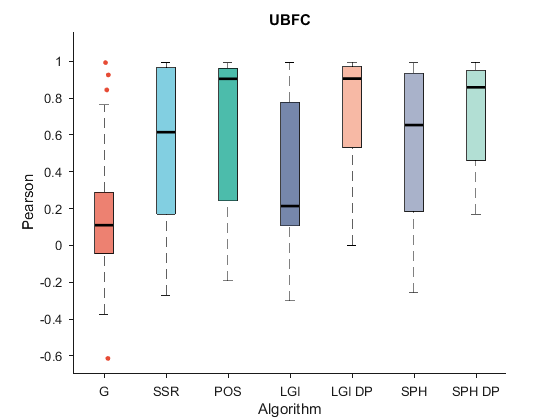}
    \caption{Pearson's correlation coefficients}
    \label{fig:ubfc_one}
  \end{subfigure}
  \begin{subfigure}[h]{0.4\textwidth}
    \includegraphics[width=\textwidth]{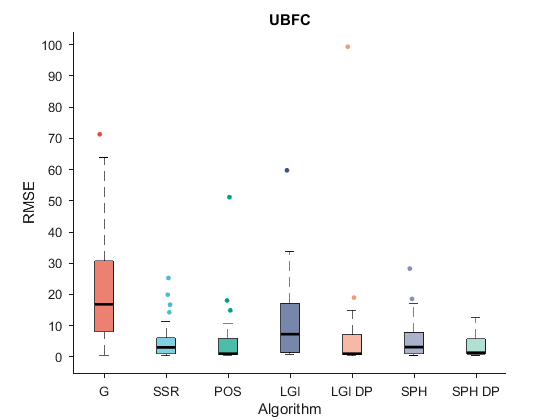}
    \caption{RMSE}
    \label{fig:ubfc_two}
  \end{subfigure}
  \caption{UBFC-RPPG: Heart rate prediction accuracy}
  \end{center}
\end{figure}

\section{Discussion}
\label{sec:discussion}

We have extended the current knowledge on linear orthogonal operators in the PPGIs feature space. The resulting manifold valued representation is holding implicit properties of invariance with respect to translations of the Group acting onto the set of features. This carries major advantages over the previously prior based assumptions, both POS and LGI, since it comes with a simple change of the topology where the existence of these properties are guaranteed  by the fundamental attributes of the space. The computational complexity of the new feature operator is supremely low and it's implementation fairly easy. The comparison against the most popular representatives of this algorithmic family succeeded with a quite promising strength of prediction accuracy. Since the approach is reflecting a fully closed form solution regarding the genuine problem statement no nasty tuning of parameters is necessary, it's operating free of parameters. The major limitation of the operator in its current form is the restriction of invariance with respect to the  group of translations. Since we have presented the proof that the intensity of pixels doesn't contribute to the periodic characteristics of skin perfusion, it follows that is indeed a question of the wavelength.

\begin{figure} [H]
\begin{center}
  \begin{subfigure}[H]{0.4\textwidth}
    \includegraphics[width=\textwidth]{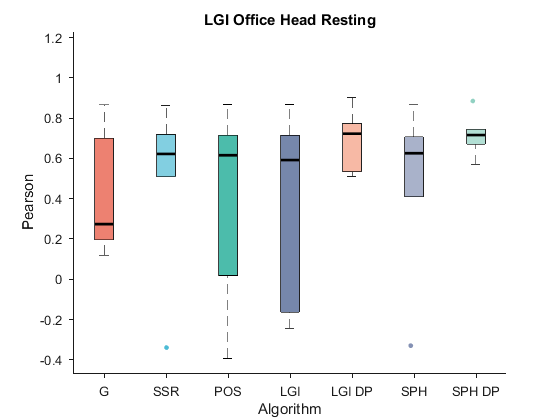}
    \caption{Session 1: Pearson's correlation coefficients}
    \label{fig:lgi_11}
  \end{subfigure}
  \begin{subfigure}[H]{0.4\textwidth}
    \includegraphics[width=\textwidth]{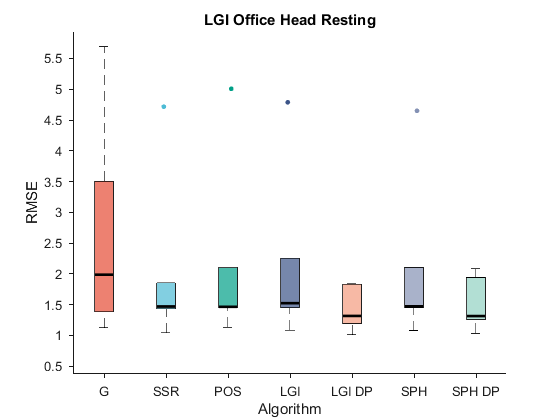}
    \caption{Session 1: RMSE}
    \label{fig:lgi_12}
  \end{subfigure}
  \begin{subfigure}[H]{0.4\textwidth}
    \includegraphics[width=\textwidth]{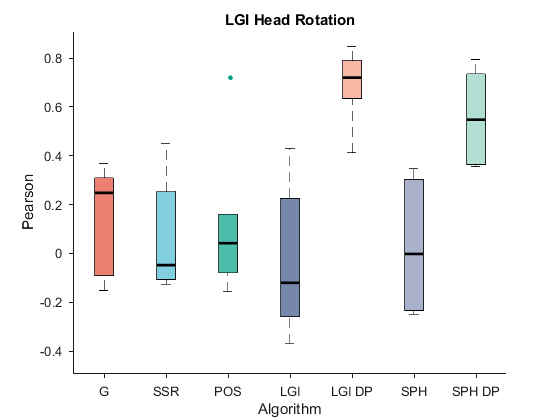}
    \caption{Session 2: Pearson's correlation coefficients}
    \label{fig:lgi_21}
  \end{subfigure}
  \begin{subfigure}[H]{0.4\textwidth}
    \includegraphics[width=\textwidth]{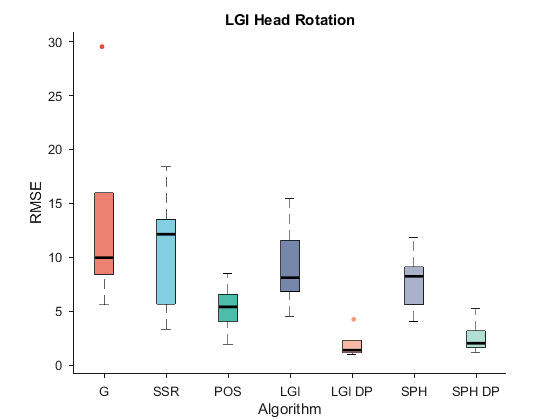}
    \caption{Session 2: RMSE}
    \label{fig:lgi_22}
  \end{subfigure}
  \begin{subfigure}[H]{0.4\textwidth}
    \includegraphics[width=\textwidth]{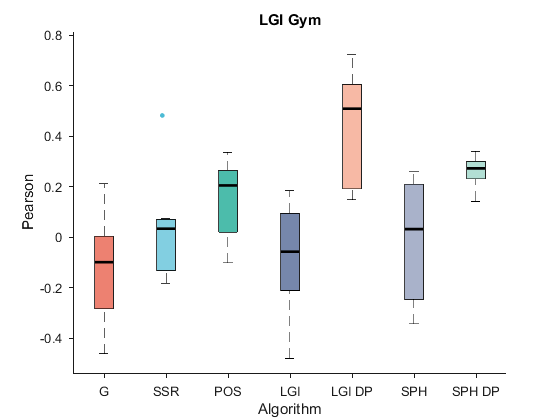}
    \caption{Session 3: Pearson's correlation coefficients}
    \label{fig:lgi_31}
  \end{subfigure}
  \begin{subfigure}[H]{0.4\textwidth}
    \includegraphics[width=\textwidth]{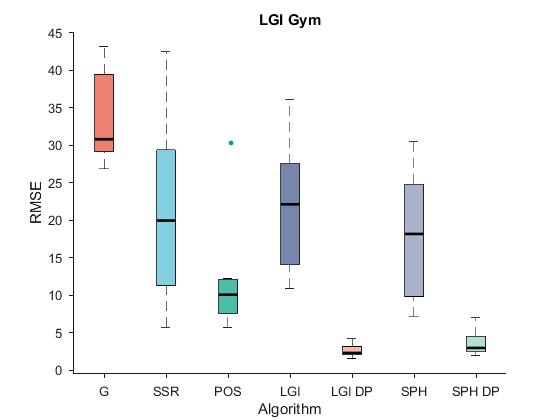}
    \caption{Session 3: RMSE}
    \label{fig:lgi_32}
  \end{subfigure}
  \begin{subfigure}[H]{0.4\textwidth}
    \includegraphics[width=\textwidth]{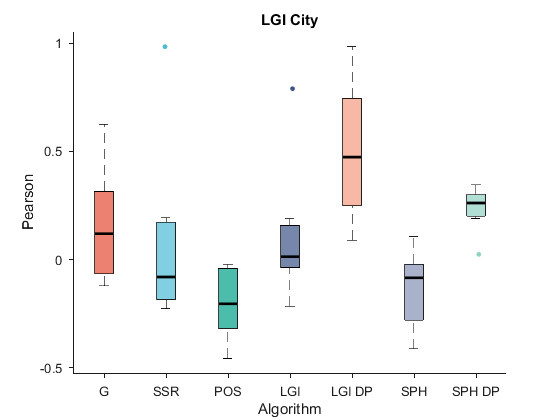}
    \caption{Session 4: Pearson's correlation coefficients}
    \label{fig:lgi_41}
  \end{subfigure}
  \begin{subfigure}[H]{0.4\textwidth}
    \includegraphics[width=\textwidth]{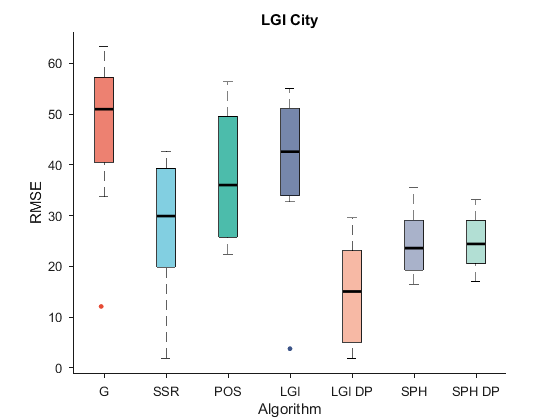}
    \caption{Session 4: RMSE}
    \label{fig:lgi_42}
  \end{subfigure}
  \caption{LGI: Heart rate prediction accuracy on different sessions. Session1: Head Resting, Session 2: Head rotation, session 3 : Bicycle ergometer, Session 4: Outdoor city talk}
  \end{center}
\end{figure}

\bibliographystyle{unsrt}
\bibliography{template}

\end{document}